\documentclass[letterpaper, 10 pt, conference]{ieeeconf}
\usepackage{amsmath,amsfonts}
\usepackage{array}
\usepackage{enumerate}
\usepackage[utf8]{inputenc}
\usepackage[T1]{fontenc}
\usepackage{amsfonts}
\usepackage{amssymb}
\usepackage{tabularx}
\makeatletter
\let\NAT@parse\undefined
\makeatother
\usepackage{hyperref}
\hypersetup{
    colorlinks=true,
    linkcolor=blue,
    filecolor=magenta,      
    urlcolor=cyan,
    }
\IEEEoverridecommandlockouts 
\usepackage{algorithm}  
\usepackage{algpseudocode}  
\usepackage{amsmath}  
\usepackage{breqn}
\usepackage[caption=false]{subfig}
\usepackage{textcomp}
\usepackage{stfloats}
\usepackage{url}
\usepackage{verbatim}
\usepackage{graphicx}
\usepackage{cite}
\usepackage{color}
\usepackage{colortbl}
\let\labelindent\relax
\usepackage{enumitem}
\usepackage{multirow}
\usepackage{diagbox}
\usepackage{booktabs}

\title{\LARGE \bf Initial Task Assignment in Multi-Human Multi-Robot Teams: \\ An Attention-enhanced Hierarchical Reinforcement Learning Approach}

\author{Ruiqi Wang$^{1}\dag$, Dezhong Zhao$^{1,2}\dag$, Arjun Gupte$^{1}$ and Byung-Cheol Min$^{1}$
\thanks{ $\dag$ Equal Contribution}
\thanks{$^{1}$SMART Laboratory, Department of Computer and Information Technology, Purdue University, West Lafayette, IN, USA. {\tt\small{[wang5357, guptea, minb]@purdue.edu}.}}
\thanks{$^{2}$College of Mechanical and Electrical Engineering, Beijing University of Chemical Technology, Beijing, China. \tt\small{DZ\_Zhao@buct.edu.cn}.}
\thanks{This paper is based on research supported by the National Science Foundation (NSF) under Grant No. IIS-1846221.}}

\begin{document}
\setlength{\abovedisplayskip}{1pt} 
\setlength{\belowdisplayskip}{1pt} 

\maketitle

\begin{abstract}
Multi-human multi-robot teams (MH-MR) obtain tremendous potential in tackling intricate and massive missions by merging distinct strengths and expertise of individual members. The inherent heterogeneity of these teams necessitates advanced initial task assignment (ITA) methods that align tasks with the intrinsic capabilities of team members from the outset. While existing reinforcement learning approaches show encouraging results, they might fall short in addressing the nuances of long-horizon ITA problems, particularly in settings with large-scale MH-MR teams or multifaceted tasks. To bridge this gap, we propose an attention-enhanced hierarchical reinforcement learning approach that decomposes the complex ITA problem into structured sub-problems, facilitating more efficient allocations. To bolster sub-policy learning, we introduce a hierarchical cross-attribute attention (HCA) mechanism, encouraging each sub-policy within the hierarchy to discern and leverage the specific nuances in the state space that are crucial for its respective decision-making phase. Through an extensive environmental surveillance case study, we demonstrate the benefits of our model and the HCA inside. Experimental details are available at \url{https://sites.google.com/view/ita-aehrl}.
\end{abstract}


\section{Introduction}
Multi-human multi-robot (MH-MR) teams are emerging as promising assets for tackling complex and expansive missions, encompassing areas such as environmental monitoring, disaster recovery, and military operations \cite{dahiya2023survey}. By integrating the unique strengths, expertise, and capabilities of diverse humans and robots, these teams elevate adaptability, efficiency, and synergy to levels unattainable by single-agent human-robot collaborations \cite{jo2023affective}. However, this inherent heterogeneity that empowers these teams also weaves into coordination challenges. To unlock the full potential of MH-MR teams, it is crucial to optimize the allocation of tasks among their diverse members.

While task assignment problems within the realm of human-robot collaborations \cite{schmidbauer2023empirical,el2022hierarchical} and multi-robot teams \cite{paul2022learning,arif2022flexible} have been extensively studied, the context of MH-MR teams remains relatively unexplored. Most existing studies center on in-process task distribution, considering the states of humans and robots with task performance indicators \cite{mina2020adaptive,patel2020improving,wu2022task}.
These studies, however, tend to overlook the initial phase of task assignment, which necessitates a keen understanding of the intricate diversity and heterogeneity characteristic of MH-MR teams. The significance of this initial allocation cannot be overstated. By aligning tasks with the inherent strengths and capabilities of team members from the outset, teams can more rapidly achieve optimal performance. Conversely, a misaligned initial distribution can hinder the team efficiency, rendering subsequent reallocations or adjustments less effective, regardless of the efforts expended.

The objective of the initial task assignment (ITA) within an MH-MR team is to optimally allocate a collection of tasks, each with distinct specifications, to a team comprising multiple humans, each influenced by varied factors, and multiple robots, each with various characteristics, from the beginning \cite{wang2023initial}. While some studies \cite{humann2018modeling,humann2023modeling} have crafted models to emulate the intricacies of MH-MR teams, especially in military surveillance scenarios, they have not ventured into optimizing the ITA process. Bridging this void, a recent work \cite{wang2023initial} framed the ITA challenge as a contextual multi-attribute decision-making process, advocating for a reinforcement learning (RL) approach to tackle it. However, the multifaceted ITA demands of MH-MR teams naturally inflate the action space. For example, allocation decisions can encompass not only task assignments to team members but also the determination of collaboration patterns between humans and robots, as well as setting robot autonomy levels. As the team size and the complexity of tasks escalate, this action space also can expand exponentially. Consequently, their proposed RL method might fall short in the long-horizon ITA problem, particularly in contexts with large-scale MH-MR teams or intricate tasks, thereby limiting its efficiency and applicability.

Hierarchical reinforcement learning (HRL), which segments a long-horizon RL task into more manageable sub-tasks, stands out as a promising solution to this challenge \cite{pateria2021hierarchical}. However, when applied to the ITA problem, a challenge emerges: the exploration within these sub-tasks becomes constrained, leading to potentially sub-optimal sub-policies. This limitation stems from the static nature of the multi-attribute context (or state input). This context, encapsulating the inherent diversity of the MH-MR team and its tasks, remains unchanged across various hierarchical stages of the sub-policy, regardless of the sub-actions taken. Such uniformity impedes the flow of information from the upper to the lower tiers of the hierarchy. Moreover, this static environmental state renders the ITA reward one-step, complicating the extraction of sub-rewards from the primary reward for each sub-policy.

Addressing the outlined challenges, we introduce AeHRL: an \textbf{A}ttention-\textbf{e}nhanced \textbf{H}ierarchical \textbf{R}einforcement \textbf{L}earning framework for initial task assignment within MH-MR teams. Our approach decomposes the potentially complex, long-horizon ITA problem into a series of structured sub-problems via HRL. This decomposition aims to enhance the precision and resilience of task assignments, especially in contexts marked by large-scale MH-MR teams and multifaceted tasks. To ensure comprehensive exploration within each sub-policy space, we introduce a hierarchical cross-attribute attention mechanism (HCA). This mechanism tends to adeptly capture specific dependencies across the attributes of humans, robots, and tasks in the state space, tailored for each allocation sub-policy. Furthermore, it can facilitate hierarchical information flow from upper to lower levels by incorporating prior action embeddings into the cross-attribute attention process. The primary contributions of our work include: 

\begin{itemize}[leftmargin=*]
    \item We introduce a hierarchical reinforcement learning approach for the initial task assignment within MH-MR teams. This structure enhances efficient ITA policy learning, particularly in large-scale team settings and intricate environments.
    \item  For efficient exploration of sub-policies, we introduce a hierarchical cross-attribute attention mechanism. This design empowers each sub-policy in the ITA hierarchy to astutely recognize and utilize distinct nuances in the state space, pivotal for its particular decision-making phase.
    \item  We conducted an extensive case study on a massive environmental surveillance task to demonstrate the efficiency of the proposed AeHRL and the HCA.
\end{itemize}

\section{Problem Formulation}
\label{PF}
Hierarchical reinforcement learning aims to break down a complicated and large RL task into a hierarchy of simpler sub-tasks. Each of these sub-tasks is then addressed by training a sub-policy using conventional RL techniques \cite{pateria2021hierarchical}. \textit{Option} framework \cite{sutton1999between} is a seminal structure in HRL. It introduces the concept of temporally extended actions, or options, which can be invoked at different decision epochs by a predetermined higher-level policy.

\begin{figure}[t]
\centering
\includegraphics[width=1\columnwidth]{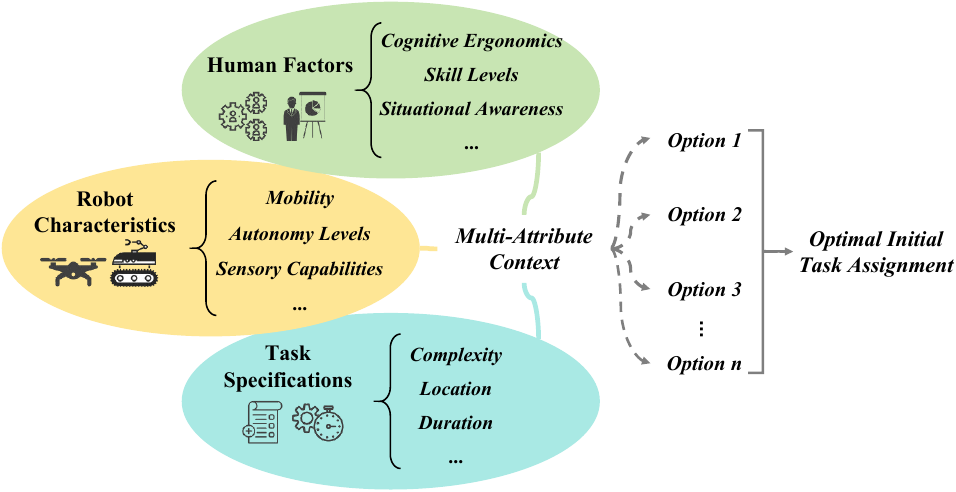}
\vspace{-17pt}
\caption{A diagrammatic representation of the formulated hierarchical contextual multi-attribute decision-making process.}
\vspace{-15pt}
\label{fig:concept}
\end{figure}

Incorporating the \textit{option} framework to the problem formulation in \cite{wang2023initial}, we present the hierarchical contextual multi-attribute decision-making process (HCMADP) formulated for the initial task assignment problem within an MH-MR team. As depicted in Fig. \ref{fig:concept}, this involves devising a hierarchy of sub-allocation decisions, termed allocation options, in response to a multi-attribute context that represents the intrinsic heterogeneity of the team and the unique specifications of tasks. Formally, the HCMADP can be characterized as a tuple $<\mathcal{J}, \mathcal{M}, \mathcal{C}, \Omega, \mathcal{R}, \mathcal{T}>$, where:

\begin{itemize}[leftmargin=*]
    \item  $\mathcal{J}$ := \{${{ta}_1}$, \dots, ${{ta}_j}$\} presents a job that the MH-MR team is tasked with. This mission is subdivided into a finite collection of $j$ tasks, each with unique specifications in terms of various aspects such as complexity, spatial attributes, and expected duration.
    \item  $\mathcal{M}$ := \{${{hm}_1}$, \dots, ${{hm}_k}$, ${{rm}_1}$, \dots, ${{rm}_i}$\} denotes an MH-MR team consisting of $k$ human members, each influenced differently by varied factors like cognitive ergonomics, operational skills, and situational awareness, and $i$ robot members, each with distinct characteristics like mobility and sensory capabilities.
    \item  $\mathcal{C}$ := \{${c_{hm}}$ $\times$ ${c_{rm}}$ $\times$ ${c_{ta}}$\} represents a multi-attribute context observed from the team and its designated mission, encompassing collective factors of the $k$ human members assessed across $h$ domains, ${c_{hm}}$ := \{\{${hf^{1}_1}$, \dots, ${hf^{1}_h}$\} $\times$ \dots $\times$ \{${hf^{k}_1}$, \dots, ${hf^{k}_h}$\}\}, combined characteristics of the $i$ robot members quantified in $r$ aspects, ${c_{rm}}$ := \{\{${rc^{1}_1}$, \dots, ${rc^{1}_r}$\} $\times$ \dots $\times$ \{${rc^{i}_1}$, \dots, ${rc^{i}_r}$\}\}, and joint specifications of the $j$ tasks measured in $w$ dimensions, ${c_{ta}}$ := \{\{${ts^{1}_1}$, \dots, ${ts^{1}_w}$\} $\times$ \dots $\times$ \{${ts^{j}_1}$, \dots, ${ts^{j}_w}$\}\}.
    \item  $\Omega$ := \{${\omega_1}$ $\times$ \dots $\times$ ${\omega_n}$\} is the joint initial task assignment decision, consisting of a hierarchy of $n$ allocation options, each is characterized as a tuple $<\pi_\omega, \tau_\omega>$. $\pi_\omega$ represents the intra-option policy for a given option, specifying the sub-allocation strategy $a_\omega$. Meanwhile, $\tau_\omega$ establishes the criteria for both initiating and terminating an option within the decision-making hierarchy, which is influenced by the overarching policy $\pi_\Omega$ that can be tailored based on the preliminary requirements or insights of the mission the MH-MR team is set to undertake. This high-level policy sends a top-down signal that determines the specific internal decision-making step $\tau$, where an option is executed as $a_{\tau} = a_\omega$ within the broader time step $t$. Namely, an option is executed only if the signal $\tau_\omega$ from the $\pi_\Omega$ remains consistent; otherwise, the process transitions to the next available option in the hierarchy. 
    \item  $\mathcal{R}$ := $Rf\left({r_t}|\left(\mathcal{C}_t, \Omega_t\right)\right)$ represents the reward function that yields a reward $r_t$ once the MH-MR team accomplishes the job by implementing the joint initial task assignment $\Omega_t$ within the multi-attribute context $\mathcal{C}_t$. This reward is determined by aggregating the performance outcomes of completing the $j$ individual tasks.
    \item  $\mathcal{T}$ := $\mathcal{P}\left({\mathcal{C}}^{\prime} \mid {\mathcal{C}} \right)$ represents the probabilistic transition of the observed multi-attribute context.
\end{itemize}


The HCMADP can be regarded as a continual contest between an agent and an opponent. At every iteration $t$, the opponent chooses a context ${\mathcal{C}_t}$ depending on a random transition function $\mathcal{T}$, and contexts can be revisited multiple times. In reaction, the agent picks a joint initial task assignment action $\Omega_t$, encompassing a hierarchy of $n$ options. The objective of the agent is to learn a hierarchical group of optimal intra-option policies $\pi^*_{\omega_n}: \mathbf{c}_{\omega} \mapsto \mathbf{a}_{\omega}$ that jointly maximizes the expected sum reward $\mathbb{E} [\sum_{s=1}^t Rf\left({r_t}|\left(\mathcal{C}_t, \Omega_t\right)\right)]$, reflecting the team performance under different contexts.

This formulation captures the core intricacies of the ITA challenge in the context of multi-human multi-robot teams, enabling the segmentation of the extensive and high-dimensional ITA action space. However, it is important to highlight the distinctions between the HCMADP and the traditional \textit{option} framework: 1) In the HCMADP, the multi-attribute context (or state) encapsulates the intrinsic heterogeneity of the MH-MR team and tasks. Contrary to the Markov decision process in the \textit{option} framework, this context is not altered by each internal decision-making step $\tau$. Instead, the transitions in context are solely influenced by the notion of the opponent. As a result, the flow of information from the higher levels to the lower ones in the option hierarchy is not accessible, and each intra-option policy operates somewhat in isolation, without the benefit of guidance or feedback from higher-level decisions; 2) The reward in HCMADP, represented by $r_t$ at each broader time step, is immediate or one-step. It can be calculated only after the team completes all tasks with the joint initial task assignment action $\Omega$. This makes it challenging to allocate a portion of the overall reward to each intra-option policy or to devise a sub-reward structure. These two distinctions in HCMADP introduce unique challenges when optimizing individual intra-option policies.

\vspace{-5pt}
\section{Methodology}
\begin{figure*}[!t]
\centering
\includegraphics[width=\linewidth]{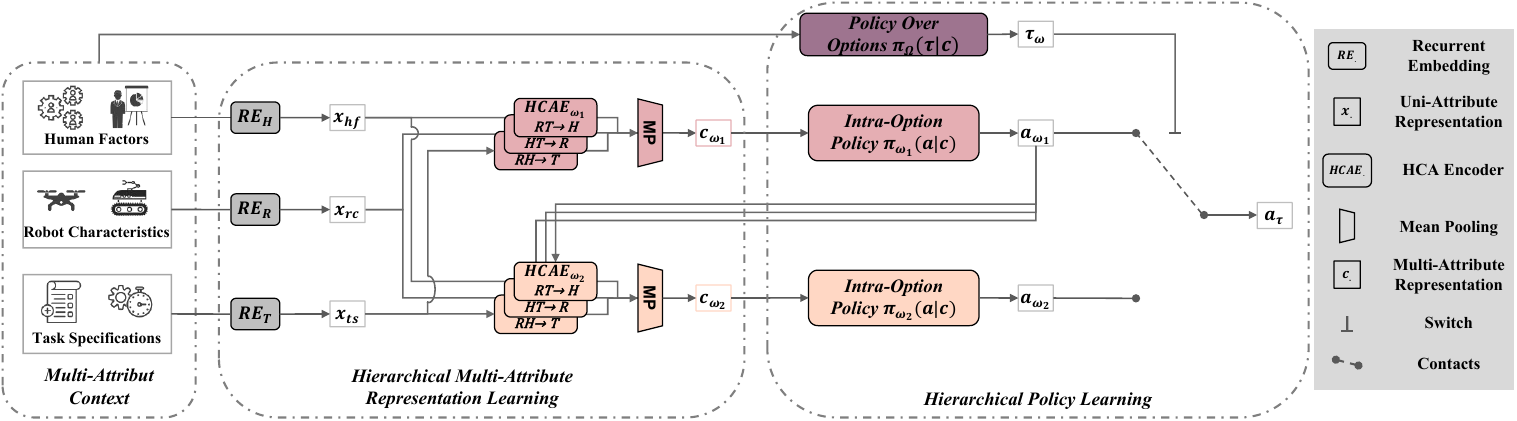}
\vspace{-18pt}
\caption{An illustration of the proposed AeHRL framework with an example of a two-level option hierarchy. It takes a multi-attribute context, which captures the heterogeneity of the MH-MR team and tasks, and learns a hierarchy of optimal initial allocation options as the output. The hierarchical execution of options is illustrated by the \textit{switch} over \textit{contacts}.}
\vspace{-15pt}
\label{fig:framwork}
\end{figure*}
\subsection{Architecture Overview}
Our aim is to address the formulated HCMADP and its associated sub-policy optimization challenge. To this end, we introduce the AeHRL, an attention-enhanced hierarchical reinforcement learning framework. Fig. \ref{fig:framwork} provides a visual representation of the AeHRL, showcasing an example of a two-level option hierarchy. However, it should be mentioned that this framework can be adapted to accommodate more options depending on the ITA task complexity. The foundational element in the proposed framework is the hierarchical multi-attribute representation learning, crafted to facilitate efficient exploration and comprehension of each option and its intra-option policy. Specifically, we introduce a hierarchical cross-attribute attention mechanism. This mechanism allows each option to derive an option-specific multi-attribute context representation, pinpointing dependencies and correlations across various attribute domains and their sub-attributes that are essential and beneficial for the current ITA decision-making sub-phase. Furthermore, this mechanism ensures a flow of decision-making information in the hierarchy.

\subsection{Recurrent Embedding}
\label{RE}
Assume an MH-MR team comprised of $k$ humans and $i$ robots, collaborating to complete a job segmented into $j$ tasks. Following the HCMADP formulation, we can obtain three uni-attribute contexts: the human factor context $c_{hf} \in \mathbb{R}^{{k},{h}}$, robot characteristic context $c_{rc} \in \mathbb{R}^{{i},{r}}$, and task specification context $c_{ts} \in \mathbb{R}^{{j},{w}}$. Subsequently, each uni-attribute context vector is transformed non-linearly using fully connected layers complemented by the Leaky ReLU activation function. This transformed vector is then passed through an LSTM cell with $d$ hidden units, extracting the hidden state from the final time step. This state serves as the uni-attribute representation for each attribute domain. Formally, this recurrent embedding procedure is expressed as:
\begin{equation}
\begin{split}
{x}_{hf}&=\dot{c}_{hf}^k=\operatorname{LSTM}\Bigl(\dot{c}_{hf}^{k-1}, {nt}\bigl({c}_{hf}^k\bigr)\Bigr)\\
{x}_{rc}&=\dot{c}_{rc}^i=\operatorname{LSTM}\Bigl(\dot{c}_{rc}^{i-1}, {nt}\bigl({c}_{rc}^i\bigr)\Bigr)\\
{x}_{ts}&=\dot{c}_{ts}^j=\operatorname{LSTM}\Bigl(\dot{c}_{ts}^{j-1}, {nt}\bigl({c}_{ts}^j\bigr)\Bigr)
\end{split}
\label{eq:1}
\end{equation}
\noindent where $\dot{c}_{hf}^k$, $\dot{c}_{rc}^i$, and $\dot{c}_{ts}^j$ represent the extracted hidden states at the final time steps for each uni-attribute, specifically for human factors $k$, robot characteristics $i$, and task specifications $j$, respectively. The term $nt$ denotes the nonlinear transformation applied.

The output uni-attribute representations, ${x}_{hf} \in \mathbb{R}^{{k},{d}}$, ${x}_{rc} \in \mathbb{R}^{{i},{d}}$ and ${x}_{ts} \in \mathbb{R}^{{j},{d}}$, all possess the same feature dimension $d$. This uniformity ensures mathematical compatibility for following hierarchical cross-attribute computations. More importantly, each representation can encode the low-level spatial features, which is beneficial due to the order-dependent nature of the ITA decision.



\subsection{Hierarchical Cross-Attribute Attention}
\label{SR}
\begin{figure}[t]
\centering
\includegraphics[width=0.95\columnwidth]{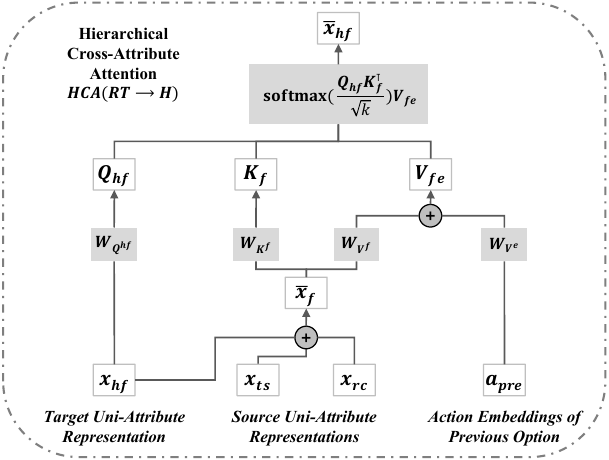}
\vspace{-5pt}
\caption{A visual representation of the hierarchical cross-attribute attention mechanism with the human factor representations ${x}_{hf}$ as a representative target uni-attribute.}
\vspace{-15pt}
\label{fig:CA}
\end{figure}

The objective of the proposed hierarchical cross-attribute attention (HCA) mechanism is to derive an option-conditioned multi-attribute context representation. To achieve this, for each option, three HCA encoders are utilized before the intra-option policy network. Each encoder refines a target uni-attribute representation by integrating complementary insights from the other two source representations, guided by an option-specific prior which are embeddings of the allocation decision of the upper-level option. In the following, we delve into the working procedure of the HCA within a specific option, using the human factor representations ${x}_{hf}$ as a representative target uni-attribute. This process is visually illustrated in Fig. \ref{fig:CA}. 

The three uni-attribute representations are first concatenated, resulting in a low-level fusion representation $\overline{x}_{f} \in \mathbb{R}^{{f},{d}}$ with $f$ being the sum of $k$, $i$, and $j$. We define the human factor Query $Q_{hf}$, fusion Key $K_{f}$, and option-specific Value $V_{fe}$ as:
\begin{equation}
\label{Cross_qkv}
\begin{aligned}
Q_{hf} &= {x}_{hf} \cdot W_{Q^{hf}}\\
K_{f} &= \overline{x}_{f} \cdot W_{K^{f}}\\
V_{fe} &= \overline{x}_{f} \cdot W_{V^{f}} \oplus {a}_{pre} \cdot W_{V^{e}}
\end{aligned}
\vspace{5pt}
\end{equation}
\noindent where $W_{Q^{hf}} \in \mathbb{R}^{d, q}$, $W_{K^{f}} \in \mathbb{R}^{d, k}$, $W_{V^{f}} \in \mathbb{R}^{d, v}$ and $W_{V^{e}} \in \mathbb{R}^{d, v}$ represent four sets of trainable weights, $\oplus$ presents concatenation operations, and ${a}_{pre}$ denotes the embeddings of the action sequence output generated by the option in the upper-level. 

Consequently, the HCA process (\textit{RT}$\rightarrow$\textit{H}) that enhances the target human factor uni-attribute by leveraging the other two source uni-attributes can be formulated as
\vspace{5pt}
\begin{equation}
\label{eq_ca}
\begin{split}
\overline{x}_{hf} &= HCA(RT \rightarrow H)\\
&= Att (Q_{hf}, K_{f}, V_{fe})\\
&= \operatorname{softmax}\left(\frac{Q_{hf} \cdot {K_{f}}^\top}{\sqrt{k}}\right){V_{fe}}.
\end{split}
\end{equation}
\noindent where $\overline{x}_{hf} \in \mathbb{R}^{k, v}$ presents an intermediate attention output.

In contrast to the cross-attention mechanism \cite{wang2023initial}, the HCA incorporates the action embeddings from the preceding option, ${a}_{pre}$, directly into the Value, while deliberately leaving the dictionary, i.e., Query and Key, unaltered. This design choice is rooted in the fundamental roles of these components as described in \cite{vaswani2017attention}: while the Query and Key collaboratively determine which elements of the source attributes are latently relevant for the target attribute, the Value essentially dictates which specific content to be emphasized in the attention output. Similar to the high-level prior information in Bayesian Inference \cite{lee2002top}, the action embeddings act as an attention beacon. This top-down cue directs the target uni-attribute to focus on the option-relevant complementary information from other source attributes, particularly through the learning of the weight $W_{V^{e}}$. Consequently, each option can derive a tailored multi-attribute context representation, thereby streamlining the exploration process of each intra-option policy. 

Furthermore, by integrating the decision-making nuances of preceding options via these action embeddings, the current option can be informed by and in harmony with prior option decisions, bolstering the consistency and efficacy of the entire hierarchical reinforcement learning framework. However, for the initial option, we omit the concatenation of previous action embeddings, operating under the assumption that decisions at this level are made based purely on the current context, uninfluenced by preceding options.

The HCA procedure in Eq. \ref{eq_ca} can be executed concurrently multiple times, referred to as multi-head HCA. Then the learned option-specific representation of human factor denoted as $c_{hf}\in \mathbb{R}^{k, d}$ can be computed as:
\vspace{5pt}
\begin{equation}
\label{chf}
c_{hf}=\mathcal{L}\left(\mathcal{F}\left(\overline{x}_{hf}\right)+{x}_{hf}\right)
\vspace{5pt}
\end{equation}
\noindent where $\mathcal{L}$ signifies the layer normalization process, and $\mathcal{F}$ corresponds to the feed-forward layers.

Following the above procedures, we can also derive the option-specific uni-attribute representations of robot characteristics, denoted as $c_{rc}$, and task specifications, represented as $c_{ts}$. Then the final context representation $c_{\omega_n}$ for a certain option $\omega_n$ can be obtained as:
\vspace{5pt}
\begin{equation}
c_{\omega_n}=\mathcal{M}\left(c_{hf}, c_{rc}, c_{ts}\right)
\vspace{5pt}
\end{equation}
\noindent where $\mathcal{M}$ represents the mean pooling operation, employed to mitigate the computational overhead potentially arising from the size of the MH-MR team or the tasks.

\subsection{Hierarchical Policy Learning}
After deriving the option-specific context representation $c_{\omega_n}$ for each option, the next step is to train a hierarchy of intra-option policies $\pi_{\omega}(a_{\omega}|c_{\omega})$. Each of these policy networks is a Gated Recurrent Unit (GRU), selected by its proficiency in capturing the temporal dynamics inherent in the ITA decision-making process. The reward allocated to each intra-option policy is the one-step reward described in Section \ref{PF}, denoted as $r_{\omega} = r_t$. We train each intra-option policy network to maximize this reward by the Proximal Policy Optimization (PPO) algorithm \cite{schulman2017proximal}. In line with the HCMADP, the training and deployment sequence of these sub-policies follow a hierarchical structure, steered by the pre-established policy over options $\pi_\Omega$ This policy is defined based on the specific task scenario. Importantly, this pre-defined overarching policy does not substantially limit the adaptability of our model. We validate this claim by implementing AeHRL across different configurations, with varying numbers of options, within the same task scenario, as discussed in Section \ref{caseset} and \ref{results}.

\section{Case Study and Experiments}
\label{case}

\subsection{Task Scenario and HCMADP Configuration}
\label{caseset}
We design a case study to assess the performance of our proposed AeHRL in the scenario of an extensive environmental surveillance task, mirroring real-world military or disaster recovery situations. In the designed scenario, potential hazards, like unauthorized vehicles or dangerous substances, necessitate meticulous monitoring and identification by an MH-MR team. The surveillance mission kicks off when a satellite system spots and lists various points of interest (POIs). Subsequently, the team is entrusted with two primary duties: 1) Navigating to each POI to capture images. This can be executed either through a fully autonomous operation by a robot or a collaborative control pattern, where a human operator offers essential navigation and imaging guidance; and 2) Analyzing the procured images to ascertain if a POI presents a real hazard or not. This evaluation can be performed by a human or via an object detection algorithm onboard the robot.

The setting of the multi-attribute context $\mathcal{C}$ in the HCMADP for the case study is as follows:
\begin{itemize}[leftmargin=*]
    \item Human Factors: We emphasize two key elements known to impact human performance in human-robot collaborations \cite{harriott2013modeling}: cognitive ability, which gauges the resilience to fatigue and workload of an individual, and operational skill level, representing the depth of expertise or mastery in specific tasks, typically honed through training and practical experience. 
    \item Robot Characteristics: We consider two main aspects: mobility (speed) and sensory capabilities (camera quality) as manifested in two main robot categories: unmanned aerial vehicles (UAVs) and unmanned ground vehicles (UGVs). UAVs, renowned for their rapid mobility, deliver an aerial perspective of the designated POIs, while UGVs, known for their deliberate movement, furnish a detailed, ground-level view. Additionally, both these robot categories can operate under two distinct autonomy levels: full autonomy or a collaborative control mode, particularly during navigation and image acquisition tasks.
    \item Task Specifications: This includes the geographical positioning of the POIs and the inherent complexity (high, medium, and low) associated with evaluating the potential hazard at a specific POI.
\end{itemize}

Regarding the decomposition of the ITA problem, specifically the policy over options $\pi_\Omega$ that determines the hierarchy of options, our case study considers four configurations, resulting in three distinct model variations:
\begin{itemize}[leftmargin=*]
    \item $\text{AeHRL}_4$: This model is structured as a four-level option hierarchy. The initial option allocates a list of POIs to each robot, directing their navigation and image acquisition activities. The subsequent two options specify the autonomy level each robot should maintain while navigating to and capturing images at the designated POIs. Concurrently, these options also determine which human operators will co-control the robot during its navigation and image capture phases at each POI respectively. The fourth option delineates the methodology for hazard detection at every POI, deciding whether the robot performs it onboard or streams the images back for human assessment. If the latter, it then assigns the image analysis tasks to specialized human experts.
    \item $\text{AeHRL}_3$: In this variant, the second and third options of the  $\text{AeHRL}_4$ are combined into a single comprehensive option.
    \item $\text{AeHRL}_2$: This variant integrates the first and second options of the $\text{AeHRL}_3$ into a unified option.
\end{itemize}

Additionally, we characterize the reward $r_t$ in the HCMADP as the cumulative points garnered from each hazard classification task at each POI. Specifically, a correct hazard identification yields points based on its inherent complexity: 15 points for low, 25 for medium, and 35 for high complexity; an incorrect identification deducts the same points. The identification success rate is determined by the human and robot performance models elaborated in the below sections.

\subsection{Simulation Environment}
\begin{figure}[h]
\vspace{-10pt}
\centering
\includegraphics[width=0.95\columnwidth]{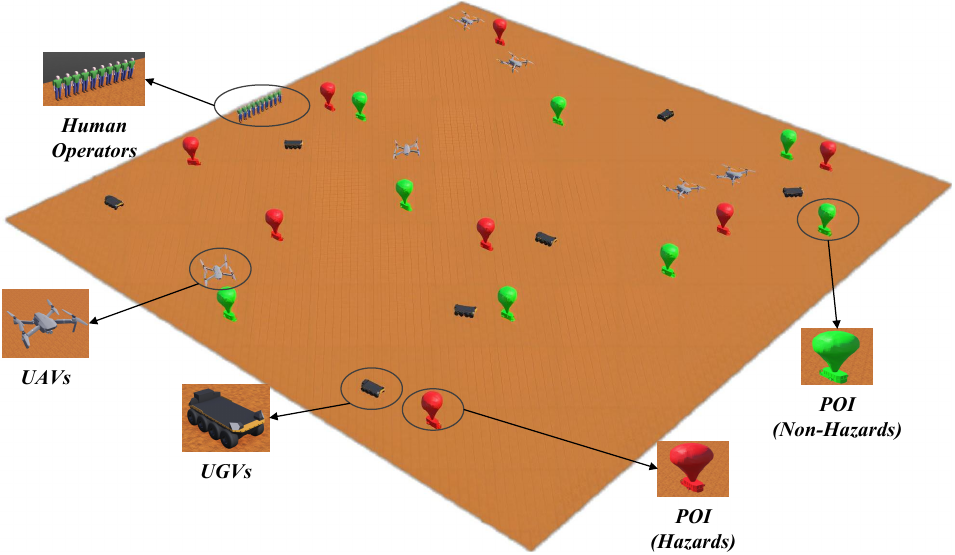}
\vspace{-5pt}
\caption{A depiction of the simulation environment. The environment scale has been magnified for enhanced clarity and visualization.}
\vspace{-5pt}
\label{fig:sim}
\end{figure}

For experiments, we employed the Webots \cite{michel2004cyberbotics} and OpenAI Gym \cite{brockman2016openai} to create a simulation environment. Inspired by configurations from previous studies \cite{wang2023initial,humann2018modeling}, we defined a 2~$km~\times$ 2~$km$ area requiring hazard surveillance by an MH-MR team, interspersed with multiple POIs. A visual representation of this environment can be seen in Fig. \ref{fig:sim}.

\subsubsection{Robot Model}
The UAV is implemented as a DJI Mavic 2 Pro drone model in the Webots while the UGV is a Clearpath Moose ground robot. Both robots employ a PID controller for navigation. The movement of a UGV involves turning in place to align with its target POI, subsequently modulating its speed based on the distance remaining. Similarly, the UAV first aligns itself at a predetermined hovering altitude, then adjusts its roll, pitch, and yaw propeller inputs to navigate accurately to the POI. The specific attributes of the UAVs and UGVs are detailed in Table \ref{tab:speed_image_quality}. Both the speed of the robots and the quality of the images they capture are influenced by the control mode. When a robot autonomously navigates to a POI, these parameters adhere to their default values. On the other hand, when operating in a shared control mode with human intervention, these parameters are modulated according to the operational expertise of an operator. Operators with low proficiency result in diminished parameters, while those with high proficiency lead to augmented parameters. For operators of medium proficiency, the parameters stay consistent. Furthermore, we posit that the probability of onboard robot hazard classification is influenced by both the quality of the captured images and the intrinsic difficulty of the hazard. The specific values for robot classification success probability $\mathcal{P}_{rc}$ under different scenarios are presented in Table \ref{tab:task}.

For each job, the robot sets off from the starting point to its designated initial POI and stays for 10 seconds to capture an image of the potential hazard. This phase can either be autonomously executed by the robot or be under shared control with a human operator. Upon image capture, the robot either undertakes onboard hazard identification or transmits the captured images back, then moves to the next unexplored POI to continue the image acquisition process. After visiting all POIs, the robot navigates back to its initial position.

\begin{table}[t]
\centering
\caption{Speed and Image Quality of two types of robots under different situations. Auto denotes operating under fully autonomous mode. W.LO, W.MO, and W.HO present operating in collaboration with human operators possessing low, medium, and high operational skill levels, respectively.}
\vspace{-5pt}
\resizebox{\linewidth}{!}{
\begin{tabular}{c ||ccc |ccc}
\toprule
 \multirow{2}{*}{\parbox{0.8cm}{\centering \textit{Robot} \\ \textit{Type} }}
 & \multicolumn{3}{c}{\textit{Speed (m/s)}} & \multicolumn{3}{c}{\textit{Quality of Captured Image}} \\
\cmidrule(lr){2-4} \cmidrule(lr){5-7}
 & W.LO & Auto / W.MO & W.HO & W.LO & Auto / W.MO & W.HO \\
\midrule
UAV & 10 & 15 & 22 & Low & Medium & Upper-Medium \\
UGV & 4 & 6 & 9 & Medium & Upper-Medium & High \\
\bottomrule
\end{tabular}
}
\label{tab:speed_image_quality}
\vspace{-5pt}
\end{table}
\begin{table}[t]
\centering
\caption{Values of $\overline{t}$ and the accurate hazard classification probability of the robot $\mathcal{P}_{rc}$ under different conditions.}
\vspace{-5pt}
\begin{tabular}{c||cc|cc|cc}
\toprule
\multirow{3}{*}{\parbox{1.2cm}{\centering \textit{Quality of} \\ \textit{Captured} \\ \textit{Images}}} & \multicolumn{6}{c}{\textit{Hazard Difficulty}} \\
\cmidrule{2-7}
 & \multicolumn{2}{c}{Easy} & \multicolumn{2}{c}{Medium} & \multicolumn{2}{c}{Hard} \\
 & $\overline{t}$ & $\mathcal{P}_{rc}$ & $\overline{t}$ & $\mathcal{P}_{rc}$ & $\overline{t}$ & $\mathcal{P}_{rc}$ \\
\midrule
Low & 45 & 0.5 & 125 & 0.3 & 245 & 0.1 \\
Medium & 35 & 0.6 & 95 & 0.4 & 195 & 0.2 \\
Upper-Medium & 25 & 0.7 & 65 & 0.5 & 145 & 0.3 \\
High & 15 & 0.8 & 35 & 0.6 & 95 & 0.4 \\
\bottomrule
\end{tabular}
\label{tab:task}
\vspace{-18pt}
\end{table}

\subsubsection{Human Model}
Grounded in empirical research on human performance in complex tasks \cite{watson2017informing}, we conceptualize the human within the MH-MR team as an event handler, sequentially handling the tasks assigned to them. In our case study, human performance in hazard classification is shaped by fatigue and workload. However, the effects of these factors can be moderated by individual cognitive abilities. Additionally, the intricacy of hazard identification, which depends on image quality and the inherent complexity of hazards, also plays a role. This can be balanced by individual operational skill levels. Formally, the likelihood of accurate hazard classification by a human is determined in a nonlinear manner \cite{humann2018modeling} by these combined elements as:
\vspace{5pt}
\begin{equation}
\mathcal{P}_{hc} =\frac{1}{2}+\eta (E_f E_w) \cdot \lambda (E_d)
\label{HP}
\vspace{5pt}
\end{equation}
\noindent where  $\eta, \lambda \in (0,\frac{\sqrt{2}}{2})$ represent the adjustment weights corresponding to individual cognitive capacity and operational expertise, separately. $E_f \in (0,1]$ and $E_w$, $E_d$ $\in (0,1)$ denote the influences of fatigue, workload, and decision-making challenges, respectively.

This formulation ensures that the lowest probability for successful classification approaches 0.5, mirroring the outcome of a random binary guessing. The configurations and values of these influence factors are established based on \cite{humann2018modeling,humann2023modeling}. We recommend readers refer to these references for insight into the theoretical underpinnings and empirical validation. The influence factors regarding fatigue and workload are delineated as follows:
\vspace{5pt}
\begin{equation}
\label{humancls}
E_f(\hat{t})=\left\{\begin{array}{ccl}
1 & & 0 \leq \hat{t}<1 \\
-0.3 \hat{t}+1.3 & & 1 \leq \hat{t} \leq 4
\end{array}\right.
\end{equation}
\begin{equation}
\begin{small}
E_d(u)=\left\{\begin{array}{ccl}
-2.47 u^2+2.22 u+0.5 & & 0 \leq u<0.45\\
1 && 0.45 \leq u<0.65\\
-4.08 u^2+5.31 u-0.724 && 0.65 \leq u \leq 1.0
\end{array}\right.
\end{small}
\vspace{5pt}
\end{equation}
\noindent where $\hat{t}$ represents the working duration of a human agent measured in hours, and $u$ signifies the utilization of a human agent, defined as the proportion of time the operator is actively engaged in tasks over a recent 5-minute interval.


Furthermore, the factor representing task complexity can be described using a sigmoid pattern, as outlined in \cite{pew1969speed}:
\vspace{5pt}
\begin{equation}
F_s(\overline{t}) = \frac{1}{1+e^{0.04(\overline{t}-180)}}
\vspace{5pt}
\end{equation}
\noindent where $\overline{t}$, quantified as the minimum duration (in seconds) necessary for task completion, presents a metric for task complexity. The value of  $\overline{t}$, influenced by the visual quality and hazard difficulty, is described in Table \ref{tab:task}.

The values of $\eta$, adjusting for performance dips due to fatigue and workload, and $\lambda$, compensating for performance reductions stemming from task complexity, are defined as:
\begin{equation}
\eta = sin(h_c); \lambda = sin(h_s)
\end{equation}
\noindent $h_c$ and $h_s$, both within the range $(0,\frac{\pi}{4})$, represent individual cognitive capability and operational expertise, respectively. 
We categorize $h_c$ and $h_s$ into three tiers: 1) Low, falling within $(0,\frac{\pi}{12})$; 2) Medium, spanning the interval $[\frac{\pi}{12},\frac{\pi}{6}]$; and 3) High, situated in the range $(\frac{\pi}{6},\frac{\pi}{4})$. The exact value for either $h_c$ or $h_s$ is randomly selected for each context in experiments.

\begin{figure*}[t]
    \centering
    \subfloat[\label{a}]{\includegraphics[width=0.68\columnwidth]{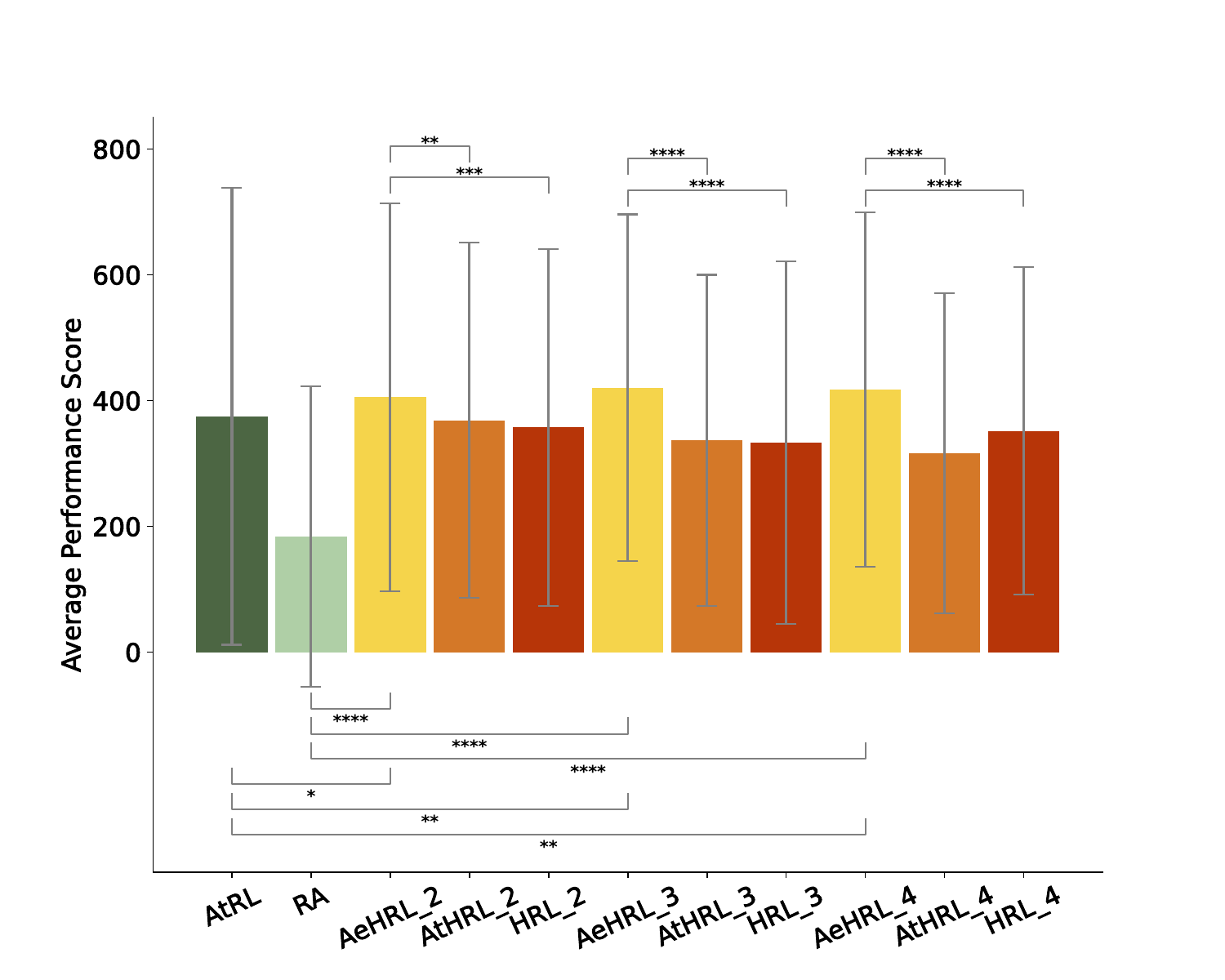}}
    \subfloat[\label{b}]{\includegraphics[width=0.68\columnwidth]{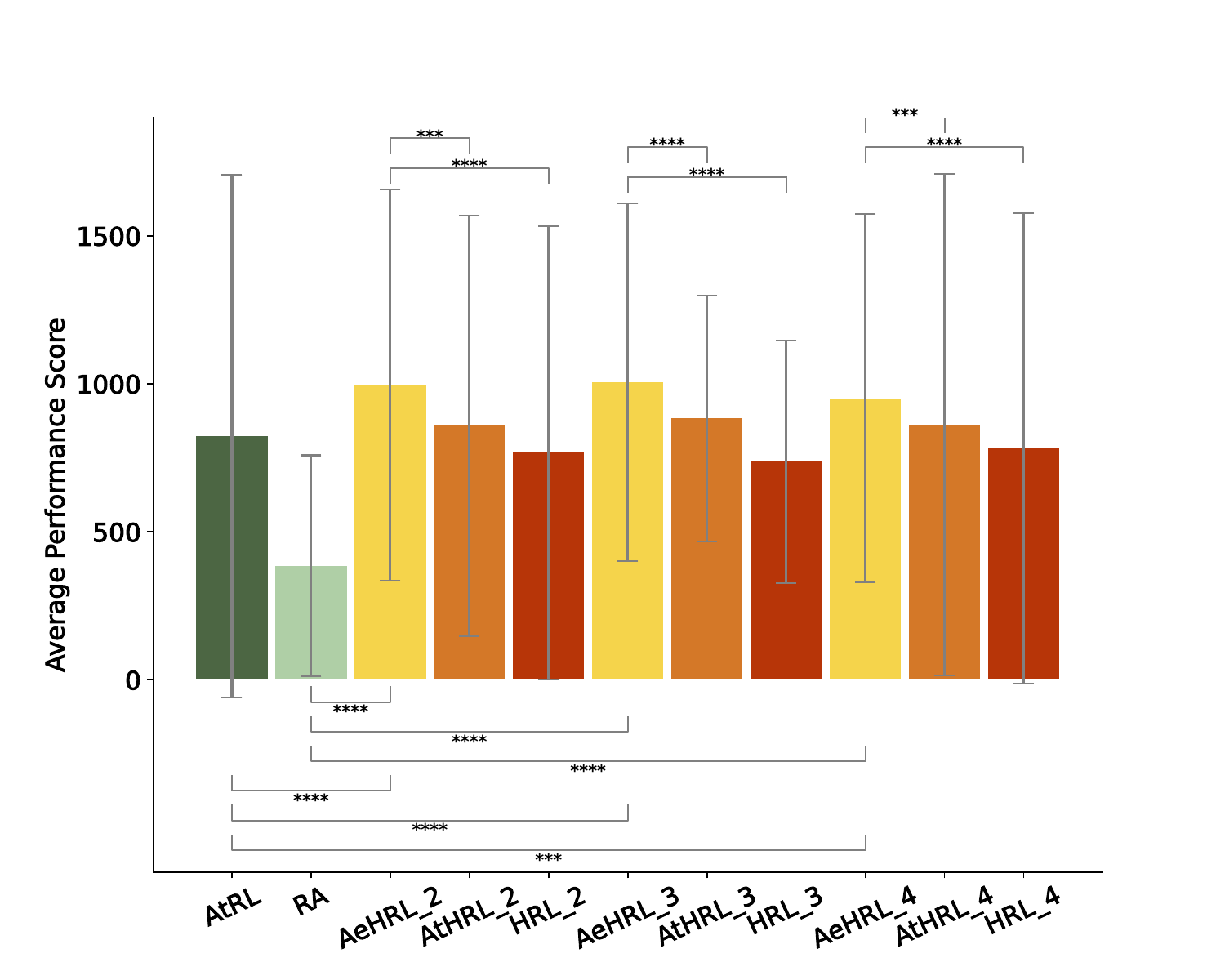}}
    \subfloat[\label{c}]{\includegraphics[width=0.68\columnwidth]{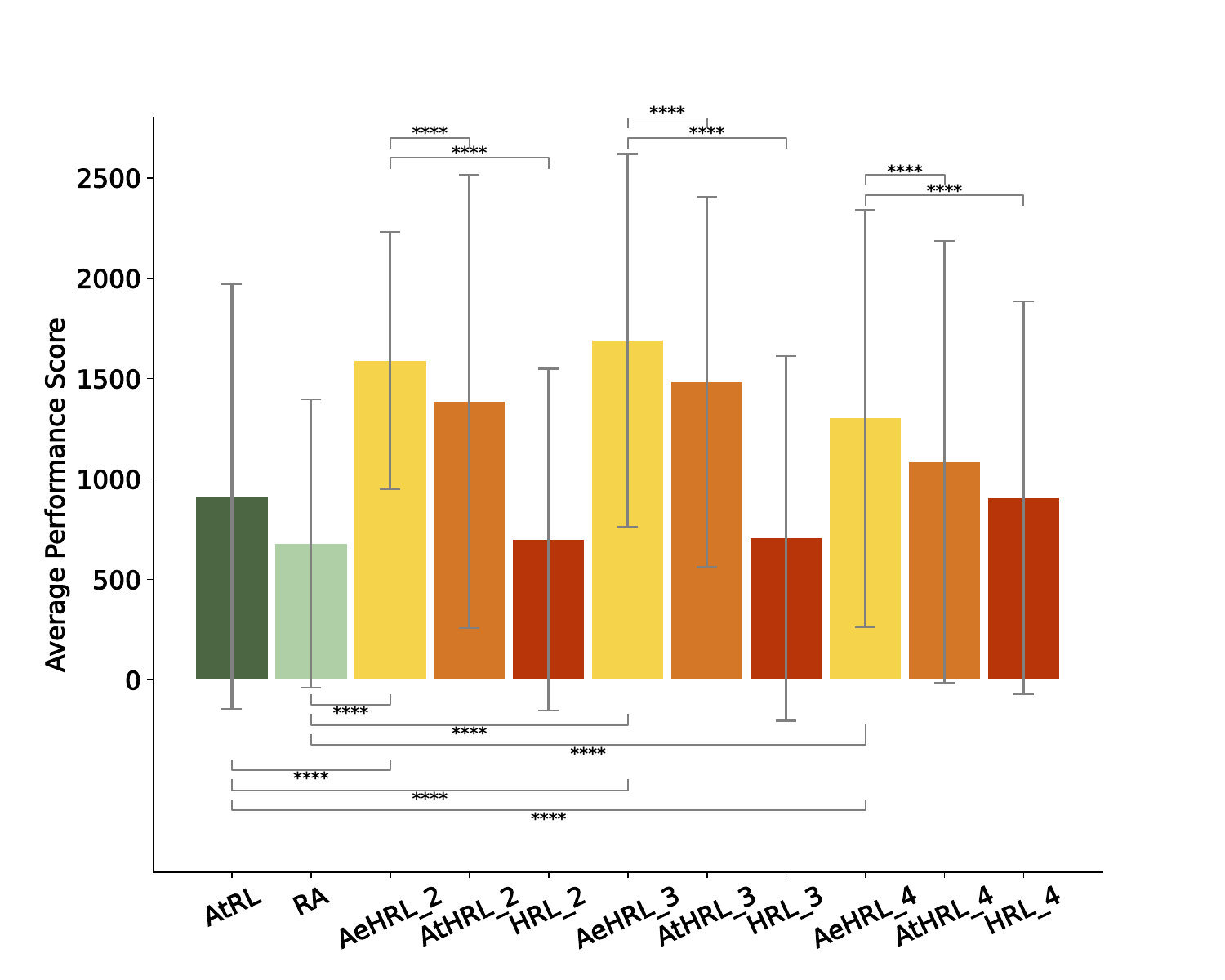}}
    \caption{Average performance scores for each model across three settings: (a) 5 humans, 7 robots, 50 POIs; (b) 10 humans, 12 robots, 100 POIs; and (c) 20 humans, 24 robots, 200 POIs. T-test results are denoted as: $*$: $p<0.1$, $**$: $p<0.05$, $***$: $p<0.01$, and $****$: $p<0.001$.}
    \label{result}
    \vspace{-10pt}
\end{figure*}

\vspace{-5pt}
\subsection{Experiments and Results}



\subsubsection{Baselines and Ablation Models}
We compared our three AeHRL model variations described in Section \ref{caseset} with two baselines: a model-based approach using random initial allocation and a learning-based approach, AtRL \cite{wang2023initial}, which leverages reinforcement learning combined with a cross-attribute attention mechanism. Additionally, we constructed two ablation models for each AeHRL variant: $\text{HRL}_{4-2}$, which omits the hierarchical multi-attribute representation learning component from AeHRLs, and $\text{AtHRL}_{4-2}$, which excludes the action embeddings from the proposed hierarchical cross-attribute attention mechanism. All RL-based models utilized identical PPO configurations for training on an NVIDIA Tesla V100 GPU.

\subsubsection{Evaluation}
We set three settings for evaluation: (a) 5 humans, 7 robots, and 50 POIs;  (b) 10 humans, 12 robots, and 100 POIs; and (c) 20 humans, 24 robots, and 200 POIs. For each setting, we test each model over 500 distinct and untrained scenarios, each characterized by unique attributes of humans, robots, and tasks. The evaluation criterion is the average performance score, representing the mean hazard classification score attained for each setup. This score is shaped by both the task specifics, particularly the hazard difficulty level at each POI, and the collective performance of the team, specifically the success rate of threat classification. Given the nature of these influencing factors, the score can exhibit significant variability. The pinnacle of performance is reached when every POI is characterized by high hazard difficulty levels and every classification task is successful, while the lowest score is recorded when all these classification tasks are unsuccessful. Furthermore, we conducted two-sample t-tests to evaluate the statistical significance when comparing each variant of AeHRL with its respective ablation models and the baseline approaches.


\begin{figure}[t]
\centering
\includegraphics[width=\columnwidth]{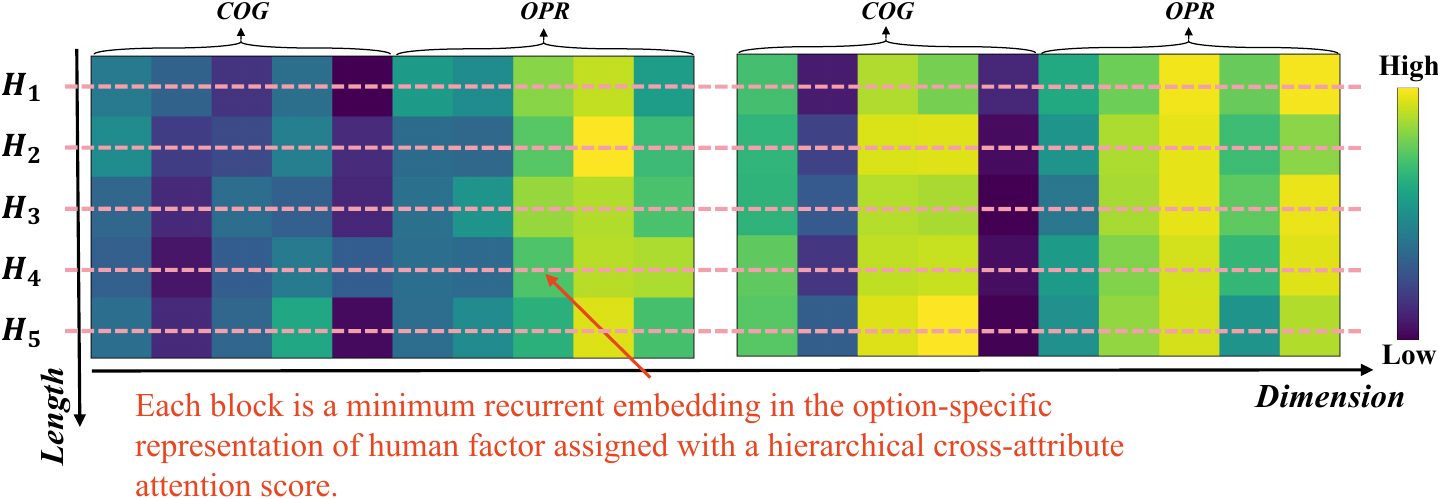}
\vspace{-18pt}
\caption{Visualizations of two option-specific human factor representations, denoted as $c_{hf}\in \mathbb{R}^{5, 10}$, learned from the hierarchical cross-attention layers of the third (left) and fourth (right) options in the $\text{AeHRL}_4$ during one training batch within the setting (a). COG and OPR denote feature embeddings of human cognitive ability and operational skill, respectively. }
\label{fig:visual}
\vspace{-18pt}
\end{figure}

\subsubsection{Results and Analysis}

\label{results}
Fig. \ref{result} showcases the comparative performance of each model across different settings in terms of the average performance score (APS) with its deviation. Generally, every variant of AeHRL consistently surpasses all baseline and ablation models across all settings on the APS. This suggests that the effectiveness of our AeHRL is maintained relatively irrespective of option configurations. Specifically, every AeHRL variant demonstrates a significantly superior performance than the model-based baseline RA in terms of APS, with a statistical significance at $p<0.001$. This is reasonable since the task assignment strategies of RA overlook the inherent heterogeneity of the MH-MR team and associated tasks. In contrast, AeHRL dynamically adapts its strategies, deriving efficient initial task distributions through continuous interactions with diverse multi-attribute contexts. This enhancement highlights the importance of awareness of the initial task allocation.

When compared to the learning-based baseline AtRL, all AeHRL variants consistently register a statistically significant performance boost on APS in settings (b) and (c) with $p<0.01$ or $p<0.001$. While the enhancement is less pronounced in setting (a), with even the $\text{AeHRL}_4$ variant reaching significance levels of $p<0.1$, the reduced APS deviation of AeHRL indicates more stable performance across varied attribute scenarios. The ascendency of AeHRL over AtRL can be attributed to its strategy of segmenting the long-horizon ITA decision-making into a series of more tractable tasks, facilitating more efficient policy exploration. On the other hand, AtRL may face challenges navigating the intricacies of a high-dimensional and expansive action space. This comparative advantage of AeHRL is accentuated in scenarios encompassing expansive MH-MR cohorts or nuanced assignments, as substantiated by empirical observations in settings (b) and (c). Collectively, these observations underscore the benefits of the decomposition of the potentially long-horizon ITA problem.


Furthermore, it is evident that each AeHRL variant markedly surpasses the performance of its ablation model HRL, as confirmed by significance levels of $p<0.01$ or $p<0.001$. We attribute this performance difference to the challenges posed by the one-step reward nature and the isolated hierarchical information flow in the ITA problem, as detailed in Section \ref{PF}. Traditional HRL methods often struggle with optimizing sub-policies under these conditions. Conversely, our AeHRL framework incorporates a hierarchical cross-attribute (HCA) mechanism, producing an option-conditioned multi-attribute context representation. This representation accentuates option-tailored dependencies and correlations in the multi-attribute context, thereby enhancing sub-policy exploration. These empirical results underscore the efficacy of the proposed HCA mechanism in the hierarchical multi-attribute representation learning module. 

Moreover, each AeHRL variant also demonstrates superior performance over its ablation model AtHRL, particularly in settings (b) and (c). We posit that this performance disparity arises from the inclusion of prior option action embeddings into the Value within the HCA in AeHRL, a component that the cross-attribute attention within AtHRL lacks. These integrations not only cultivate a more tailored option-specific contextual representation but also promote a seamless flow of hierarchical information from higher to lower tiers, ensuring a more efficient and coherent decision-making process across the hierarchy.


Additionally, to illustrate the functioning of the HCA mechanism, we present a visualization of the attention outputs. Fig. \ref{fig:visual} demonstrates two option-specific human factors representations $c_{hf}\in \mathbb{R}^{5, 10}$ learned from the hierarchical cross-attention layers of the third and fourth options in the $\text{AeHRL}_4$ under setting (a) (refer to Eqs. \ref{Cross_qkv}-\ref{chf}). A clear observation is the stable attention distribution patterns learned across individual human members from $H_1$ to $H_5$ within both options, underscoring the consistency of the HCA. Delving deeper, the attention patterns are also adaptive, revealing the option-specific nature of the HCA. For example, the HCA of the third option accentuates features emblematic of human operational skill. This aligns seamlessly with its role of assigning humans to exert shared control over robots during the image capture phase, a task intrinsically anchored in operational skill as elucidated in Table \ref{tab:speed_image_quality}. In contrast, the HCA of the fourth option magnifies attributes symptomatic of both human cognitive ability and operational expertise. This is in harmony with its task of delegating hazard categorization to humans, a responsibility shaped by the synergistic interplay of cognitive capability and operational skill as detailed in Eq. \ref{humancls}. Such observations underscore the ability of our HCA to adeptly discern and internalize option-specific contextual intricacies, thereby pinpointing pivotal dependencies vital for bolstering sub-policy learning efficacy.

\section{Conclusion}
In this study, we presented an attention-enhanced hierarchical reinforcement learning framework for the initial task assignment challenge in multi-human multi-robot teams. Central to our approach is the hierarchical cross-attribute attention mechanism, which we introduced to enhance sub-policy learning. Through a case study involving an MH-MR team engaged in a large-scale environmental surveillance task, we demonstrated that our approach can significantly improve team performance by optimally matching different tasks to individual capabilities right from the outset. Future research will explore the integration of in-process task reallocation strategies with our initial task assignment framework.


\typeout{}
\bibliography{main}
\bibliographystyle{IEEEtran}
\end{document}